\documentclass[10pt]{article}
\usepackage{ifxetex}
\usepackage{graphicx}
\usepackage{booktabs}  
\usepackage{multirow}
\usepackage{url}
\usepackage{array}

\ifxetex
    \usepackage{fontspec}
\else\ifluatex
    \usepackage{fontspec}
\else
    \usepackage{times} 
\fi\fi

\usepackage[letterpaper, margin=1in]{geometry}
\usepackage[super,comma,sort&compress]{natbib}  

\usepackage{authblk}
\usepackage{amssymb,amsmath}
\usepackage{graphicx}
\usepackage[labelfont=bf,labelsep=period]{caption}

\usepackage{sectsty}
\allsectionsfont{\rmfamily\fontsize{10}{10}\selectfont}
\subsectionfont{\rmfamily\itshape\bfseries\fontsize{10}{10}\selectfont}

\subsubsectionfont{\rmfamily\normalfont\itshape}

\usepackage{titlesec}

\titlespacing{\section}{0pt}{0.5\baselineskip}{0.5\baselineskip}
\titlespacing{\subsection}{0pt}{0.5\baselineskip}{0.5\baselineskip}

\title{\bfseries\fontsize{14}{14}\selectfont Bias Evaluation and Mitigation in Retrieval-Augmented Medical Question-Answering Systems}
\author[ ]{\bfseries\fontsize{12}{14}\selectfont Yuelyu Ji, MS\textsuperscript{1},\bfseries\fontsize{12}{14}\selectfont Hang Zhang, MS\textsuperscript{2},\bfseries\fontsize{12}{14}\selectfont Yanshan Wang, PhD, FAMIA\textsuperscript{3,4,5}}

\affil[1]{\fontsize{12}{14}\selectfont Dept. of Information Science, University of Pittsburgh, Pittsburgh, PA, USA}
\affil[2]{\fontsize{12}{14}\selectfont Intelligent Systems Program, School of Computing and Information, University of Pittsburgh, Pittsburgh, PA, USA}
\affil[3]{\fontsize{12}{14}\selectfont Dept. of Health Information Management, University of Pittsburgh, PA, USA}
\affil[4]{\fontsize{12}{14}\selectfont Clinical and Translational Science Institute, University of Pittsburgh, Pittsburgh, PA, USA}
\affil[5]{\fontsize{12}{14}\selectfont University of Pittsburgh Medical Center, Pittsburgh, PA, USA}

\date{}

\begin{document}
\maketitle

\section*{Abstract}
\emph{Medical Question-Answering (QA) systems based on Retrieval-Augmented Generation (RAG) are promising for clinical decision support due to their capability to integrate external knowledge, thus reducing inaccuracies inherent in standalone large language models (LLMs). However, these systems may unintentionally propagate biases associated with sensitive demographic attributes like race, gender, and socioeconomic factors. This study systematically evaluates demographic biases within medical RAG pipelines across multiple QA benchmarks, including MedQA, MedMCQA, MMLU, and EquityMedQA. We quantify disparities in retrieval consistency and answer correctness by generating and analyzing queries sensitive to demographic variations. We further implement and compare several bias mitigation strategies—including Chain-of-Thought reasoning, Counterfactual filtering, Adversarial prompt refinement, and Majority Vote aggregation—to address identified biases. Experimental results reveal significant demographic disparities, highlighting that Majority Vote aggregation improves accuracy and fairness metrics. Our findings underscore the critical need for explicitly fairness-aware retrieval methods and prompt engineering strategies to develop truly equitable medical QA systems.}

\section*{Introduction}

Medical question-answering (QA) systems powered by large language models (LLMs) have shown remarkable progress in knowledge-intensive tasks, promising valuable clinical decision support \cite{achiam2023gpt, verma2019lirme,chen2025improving, gao2023retrieval,ding2024enhance,deng2024composerx, li2024advances,yang2024hades,chen2024mix,li2024gptdrawerenhancingvisualsynthesis, zhao2024large, dan2024evaluation,dan2024image,202503.0300,jiang2024trajectorytrackingusingfrenet,Yang2024a}. Despite their advancements, LLMs often suffer from issues like factual inaccuracies and hallucinations, particularly critical in high-stakes domains like healthcare. Retrieval-Augmented Generation (RAG) addresses these limitations by integrating external knowledge bases to enhance factual accuracy and minimize hallucinations \cite{li2024enhancing,gautam2024overview,shrestha2024fairrag,singhal2025toward}. Specifically, RAG-based approaches retrieve relevant evidence, thus significantly improving response reliability.

However, introducing external retrieval also brings new challenges. Recent studies indicate that biases—often associated with sensitive attributes such as race, gender, or socioeconomic status—can persist or even be exacerbated within both retrieval and generation stages of RAG systems, potentially compromising fairness and reliability \cite{levra2025large, ni2025towards}.

Although prior research has investigated biases within end-to-end generative models, less attention has been directed toward disparities arising specifically during the RAG pipeline \cite{pfohl2024toolbox}. In this context, disparities can emerge when retrieval mechanisms systematically neglect or inadequately represent particular groups characterized by sensitive attributes such as race and gender. Moreover, generation models conditioned on prompts containing sensitive identity cues (e.g., \emph{“This African American patient”}) risk reinforcing stereotypes or neglecting clinically relevant nuances. Given the growing emphasis on fairness and equity in healthcare AI, understanding how sensitive identity attributes influence retrieval and generation outcomes becomes crucial.

To address this critical gap, we propose a unified framework to systematically evaluate and mitigate biases in medical RAG pipelines. Specifically, we first generate query variants that explicitly incorporate sensitive attributes (e.g., race: Caucasian, African American, Asian, Hispanic; gender: male, female, non-binary) to uncover disparities in retrieval outcomes and QA accuracy. Second, we design and compare several bias mitigation strategies, including Chain-of-Thought (COT) filtering \cite{wei2022chain} that encourages step-by-step reasoning, Counterfactual filtering that verifies consistency across varied identity contexts, and Adversarial prompts designed to prevent over-reliance on identity terms. Additionally, we introduce a Majority Vote mechanism \cite{chen2024more}, aggregating model outputs across multiple attribute-specific variants, further improving robustness and fairness.

We validate our approach across multiple diverse benchmarks—\textbf{MedQA}\cite{jin2021disease}, \textbf{MedMCQA}\cite{pal2022medmcqa}, \textbf{MMLU}\cite{hendrycks2020measuring}, and the open-ended \textbf{EquityMedQA}\cite{pfohl2024toolbox}. By leveraging fairness metrics such as Demographic Parity (DP)\cite{jiang2022generalized} and Equal Opportunity (EO)\cite{cockburn1989equal}, our experimental results quantitatively demonstrate significant biases within existing retrieval and generation processes and highlight how our proposed mitigation methods substantially reduce these disparities.

Our main contributions include:

(1) Introducing a unified evaluation framework explicitly designed to measure biases associated with sensitive identity attributes within medical RAG pipelines.

(2) Developing and empirically validating multiple bias mitigation approaches—Chain-of-Thought, Counterfactual, Adversarial prompting, and Majority Vote aggregation—to enhance fairness.

(3) Conducting extensive evaluations on standard medical QA datasets to illustrate the efficacy of our proposed methods in achieving measurable improvements in fairness metrics.

Our results highlight the importance of fairness-aware retrieval, bias-sensitive prompt engineering, and systematic aggregation methods to develop equitable medical QA systems. We publicly release our code and experimental protocols to foster reproducibility and promote further research on fairness in medical AI.\footnote{\url{https://github.com/JoyDajunSpaceCraft/EquityGuradRAG.git}}

\begin{figure}[h]
    \centering
    \fbox{\includegraphics[width=0.8\linewidth]{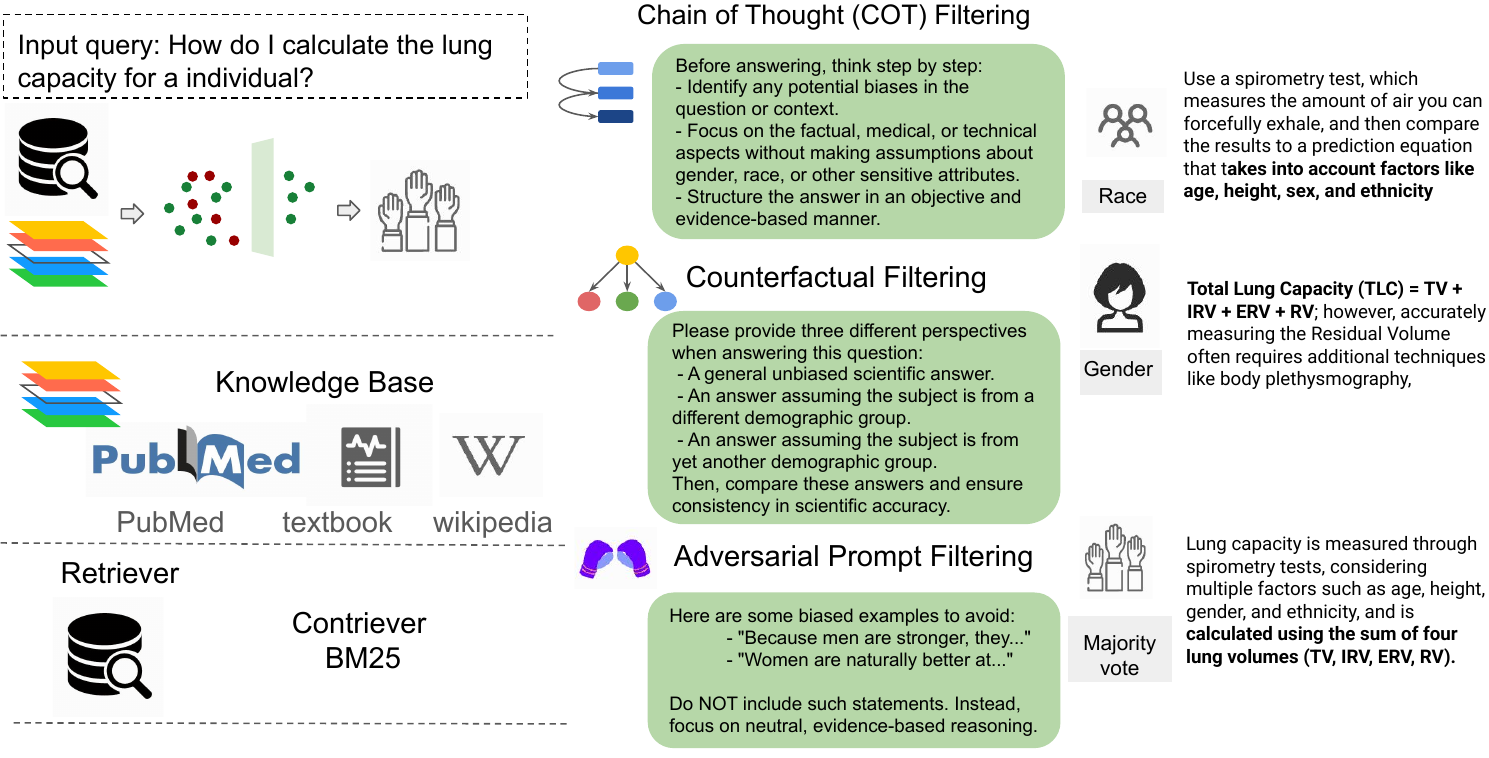}}
    \caption{Overview of our proposed bias mitigation framework for medical Retrieval-Augmented Generation (RAG), highlighting three effective filtering methods and Majority Vote aggregation.
    The system takes an input medical query, retrieves relevant documents from a knowledge base (e.g., PubMed, textbooks, Wikipedia) using a retriever (Contriever BM25), and applies multiple filtering strategies before generating an answer. Three bias mitigation techniques are incorporated: (1) \textbf{Chain of Thought (COT) Filtering}, which encourages structured, evidence-based reasoning while avoiding implicit biases; (2) \textbf{Counterfactual Filtering}, which generates responses from different demographic perspectives and ensures consistency in scientific accuracy; and (3) \textbf{Adversarial Prompt Filtering}, which identifies and avoids biased phrasing in model-generated responses. Finally, a \textbf{Majority Vote} mechanism aggregates multiple responses to mitigate potential biases further and improve answer robustness.}
   
    \label{fig:framework}
\end{figure}

\section*{Related Work}

Recent advancements in large language models (LLMs) and Retrieval-Augmented Generation (RAG) have shown significant promise in medical question-answering systems. However, ensuring the fairness and trustworthiness of these systems, especially in high-stakes medical contexts, remains a critical challenge. Ni et al.~\cite{ni2025towards} provide a comprehensive survey outlining key aspects of trustworthy RAG systems, highlighting reliability, privacy, safety, fairness, explainability, and accountability as critical dimensions of trust. They identify that fairness in RAG systems requires special attention in both the retrieval and generation stages, as biases introduced during retrieval can propagate to generation, potentially exacerbating disparities.

Several studies further explore bias in retrieval-augmented systems, specifically within healthcare contexts. Levra et al.\cite{levra2025large} emphasized the risks of demographic biases that can be inadvertently amplified through retrieval processes in medical QA systems. Similarly, Pfohl et al. \cite{pfohl2024toolbox} introduced EquityMedQA, an open-ended dataset explicitly designed to test demographic biases in medical QA models, demonstrating disparities arising due to sensitive attributes such as race and gender. Recent methods have proposed sophisticated bias mitigation strategies to address these fairness issues. Chain-of-Thought (COT) prompting~\cite{wei2022chain,yao2023tree} encourages explicit reasoning steps, potentially reducing reliance on demographic stereotypes. Counterfactual filtering~\cite{verma2024auditing} alters sensitive attributes to check model consistency and minimize discriminatory outcomes. Additionally, adversarial filtering and Majority Vote aggregation~\cite{li2024learning} have effectively reduced biases and promoted more equitable model outcomes by aggregating diverse demographic perspectives. Despite these advancements, challenges persist. Retrieval methods still sometimes select outdated or irrelevant information, as evidenced by error cases involving intersectional demographics (e.g., Asian non-binary individuals), identifying a gap in robustness and generalization capabilities. Ongoing work thus emphasizes the importance of developing trustworthy RAG systems, with comprehensive frameworks proposed to address reliability, privacy, and fairness comprehensively~\cite{ni2025towards}.

\section*{Methods}
\subsection*{Datasets Overview}

We base our experiments on four medical QA datasets that encompass multiple-choice and open-ended questions and demographic attributes such as race and gender. Specifically, \textbf{MedQA}  is a single-choice test bank covering a broad range of clinical topics, extended here with race (Caucasian, African American, Asian, Hispanic) and gender (male, female, non-binary) cues. \textbf{MedMCQA}  provides multiple-choice questions focusing on medical licensing content, similarly augmented by demographic variants. \textbf{MMLU} is a general multi-subject benchmark that includes medical categories, and we selectively incorporate demographic contexts to reveal potential group-level biases. Finally, \textbf{EquityMedQA}  serves as our open-ended QA resource, where reference solutions are compared against model-generated text via ROUGE-based evaluations by introducing sensitive attribute mentions (e.g. \emph{This African American patient...}), and each data set can comprehensively test how LLMs respond to diverse demographic backgrounds.

\begin{table}[h]
    \centering
    \caption{Dataset Overview}
    \label{tab:dataset_overview}
    \begin{tabular}{|l|c|c|c|}
        \hline
        \textbf{Dataset Name} & \textbf{Type} & \textbf{Demographics} & \textbf{Task Type} \\
        \hline
        MedQA & Single Choice  & Race, Gender & Closed QA \\
        MedMCQA & Multiple Choice & Race, Gender & Closed QA \\
        MMLU & Multiple Choice & Race, Gender & Closed QA \\
        EquityMedQA & Open-Ended & Race, Gender & Open QA \\
        \hline
    \end{tabular}
\end{table}

\subsection*{Bias Removal Filtering Methods}
We select four representative general-purpose large language models (LLMs) for integration into our Retrieval-Augmented Generation (RAG) framework, including \textbf{DeepSeek-R1-8B} , \textbf{DeepSeek-R1-70B}\cite{bi2024deepseek}, \textbf{Meta-Llama-3-8B}\cite{dubey2024llama}, and \textbf{PMC-LLaMA-13B} \cite{wu2024pmc}. These models were chosen based on their proven capabilities in general reasoning, instruction-following tasks, and suitability for adaptation into domain-specific contexts, such as medical question-answering. 

We apply four strategies to mitigate demographic bias in Retrieval-Augmented Generation (RAG). First, a \textbf{Plain (baseline)} condition involves no explicit intervention. Second, \textbf{Chain-of-Thought (COT) Filtering} \cite{wei2022chain, yao2023tree} guides models to produce step-by-step reasoning that isolates medical facts from demographic descriptions, aiming to lessen undesired influences of attributes like race or gender. Third, \textbf{Counterfactual Filtering} \cite{verma2024auditing} prompts the model with methodically altered demographic labels and checks for inconsistencies or discriminatory behaviors in the resulting answers. Lastly, \textbf{Adversarial Prompt Filtering} reformulates queries to minimize reliance on socially sensitive markers, thus preventing the model from overfitting to or amplifying potential biases. We compare each filtering method’s output regarding accuracy, retrieval patterns, and fairness measures during inference.

\subsection*{Majority Vote Aggregation}
To further combat biases that may persist for a single demographic instance, we incorporate a majority voting approach \cite{li2024learning} across multiple variants of the same question. Concretely, for each question in MedMCQA, MedQA, MMLU, and EquityMedQA, we generate distinct demographic variants by substituting race (Caucasian, African American, Asian, Hispanic) and gender (male, female, non-binary) into the query stem. For example, an original question \emph{“Which medication is recommended for a patient with chest pain?”} can yield up to 12 variants if we combine four race attributes and three gender attributes. In practice, if both race and gender are not always simultaneously varied, we produce 4 to 6 variants (e.g., only race or only gender) depending on the experiment.

Once these demographic-specific queries are formed, we prompt the model for each variant independently:
\begin{itemize}
    \item For multiple-choice tasks (A/B/C/D), each variant obtains an answer, and we select the most frequently chosen option among them as the final consensus prediction.
    \item For open-ended tasks (e.g., EquityMedQA), we gather all demographic-specific responses and compute pairwise text similarity (using sentence embeddings). We cluster the responses by similarity and pick the largest cluster as the final answer, effectively filtering out outlier or potentially biased responses.
\end{itemize}

\subsection*{Evaluation Metrics}

We employ a combination of \emph{performance} and \emph{fairness} metrics to thoroughly evaluate our Retrieval-Augmented Generation (RAG) models under both closed-form and open-ended QA tasks:

\noindent
\textbf{(1) Accuracy.} 
Each query has a discrete correct option for the \emph{closed-form QA} datasets (e.g., MedQA, MedMCQA, MMLU). The model's answer is considered correct if it matches the ground-truth choice (A/B/C/D for multiple-choice or a single correct label for single-choice). We report the percentage of questions correctly answered.

\noindent
\textbf{(2) ROUGE-L.}
For \emph{open-ended QA} such as EquityMedQA \cite{pfohl2024toolbox}, which lacks fixed answer options, we measure correctness by comparing the generated text to a reference solution using ROUGE-L. This metric quantifies the overlap of the longest common subsequence (LCS) between the model's output and the reference. A higher ROUGE-L indicates greater alignment with the reference's content, helping detect factual completeness in a free-text generation.

\noindent
\textbf{(3) Retrieval Overlap (\%)} 
We analyze the documents fetched for each demographic variant of the same query to assess whether the model retrieves consistent or demographic-specific evidence. We measure the intersection-over-union ratio of document IDs across variants as a percentage. Since open-ended generation can be more prone to hallucinations or subjective framing, retrieval overlap helps us identify when specific subgroups might receive different sources, potentially affecting fairness.

\noindent
\textbf{(4) Demographic Parity (DP).}
We define a correct model prediction as $\hat{Y}=1$. Demographic Parity checks that no demographic subgroup is comprehensively favored or disfavored in receiving correct predictions:
\[
\text{DP Disparity} \;=\; \max_{g,g'} \;\Bigl|\;P(\hat{Y}=1 \mid G=g)\;-\;P(\hat{Y}=1 \mid G=g')\Bigr|.
\]
A lower DP disparity implies the model maintains more uniform correctness rates across groups (e.g., race, gender).

\noindent
\textbf{(5) Equal Opportunity (EO).}
In medical QA, some questions are \emph{truly answerable} ($Y=1$). EO measures how fairly the model provides correct answers among these “answerable” queries. Formally,
\[
\text{EO Disparity} \;=\; \max_{g,g'} \;\Bigl|\;P(\hat{Y}=1 \;\mid\; Y=1,\;G=g) \;-\; P(\hat{Y}=1 \;\mid\;Y=1,\;G=g')\Bigr|.
\]
A lower EO disparity indicates that among all queries that \emph{can} be answered correctly, each demographic subgroup is treated relatively equally.

\textbf{Accuracy} and \textbf{ROUGE-L} capture the \emph{overall correctness} of the model on closed-form vs.\ open-ended QA. \textbf{Retrieval Overlap} helps pinpoint whether the system fetches \emph{consistent evidence} across demographic variants, shedding light on the potential retrieval-phase bias. \textbf{DP/EO} offer group-level fairness assessments, demonstrating whether model accuracy is equitably distributed or if some subgroups receive inferior answers.

\subsection*{Corpora and Retrieval Methods}
We adopt four corpora in our retrieval pipeline, each chunked into short snippets: (1) \textbf{PubMed} \cite{white2020pubmed} for biomedical abstracts,(2) \textbf{Medical Textbooks} \cite{jin2021disease} for domain-specific knowledge, and (3) \textbf{Wikipedia} \cite{glott2010wikipedia} for more general context. In addition, we combine these sources into a larger \textbf{MedCorp} if cross-domain retrieval is desired. Each snippet is indexed and retrieved via different retriever types, including a lexical approach (\textbf{BM25}) \cite{robertson2009probabilistic} and semantic encoders (Contriever) \cite{cohan2020specter, izacard2021unsupervised}. By default, we retrieve $k=15$ snippets per query; if multiple retrievers are used, we employ \emph{Reciprocal Rank Fusion (RRF)} \cite{cormack2009reciprocal} to merge results.

\section*{Results}
\subsection*{Overall Model Performance}
Table~\ref{tab:final_results} summarizes the performance of four Large Language Models evaluated across medical QA benchmarks, including closed-form datasets (\textbf{MedQA}, \textbf{MedMCQA}, \textbf{MMLU}) and an open-ended dataset (\textbf{EquityMedQA}). Five strategies are tested: \emph{Plain}, \emph{Chain-of-Thought (COT)}, \emph{Counterfactual}, \emph{Adversarial}, and a subsequent \emph{Majority Vote} step. Accuracy (\%) is reported for closed-form tasks, while ROUGE-L (\%) is used for EquityMedQA. Additionally, retrieval overlap (\%) indicates consistency in retrieved documents.

\textbf{DeepSeek-R1-70B} consistently outperforms other models, achieving up to 34\% accuracy on MedQA and 32.2\% on MMLU. Conversely, smaller models such as \textbf{Meta-Llama-3-8B} exhibit lower accuracy despite higher retrieval overlap, indicating possible mismatches between retrieval and generation components. For OpenQA (EquityMedQA), applying \emph{Majority Vote} increases ROUGE-L scores, reaching 52.0\%, identifying the importance of aggregating multiple demographic perspectives.

\begin{table*}[ht]
\centering
\caption{Comparison of five filtering approaches \emph{(Plain, COT, Counterfactual, Adversarial, Majority Vote)} across four LLMs  \textbf{(DeepSeek-R1-8B, DeepSeek-R1-70B, Meta-Llama-3-8B, PMC-LLaMA-13B)} on four medical QA datasets. 
For closed-form QA tasks (MMLU, MedQA, MedMCQA), we report \textbf{Accuracy (\%)}. 
For open-ended QA (EquityMedQA), we report \textbf{ROUGE-L (\%)}. All are shown as \emph{Score}. 
In the \emph{no-vote} scenario (Plain, COT, Counterfactual, Adversarial), we use a \emph{single demographic variant}, 
while \emph{Majority Vote} aggregates multiple variants. 
Retrieval Overlap (\%) is the intersection-over-union of documents fetched per demographic variant.}
\label{tab:final_results}
\resizebox{\textwidth}{!}{%
\begin{tabular}{|l|l|cc|cc|cc|cc|cc|}
\hline
\multirow{2}{*}{\textbf{Model}} 
& \multirow{2}{*}{\textbf{Dataset}} 
& \multicolumn{2}{c|}{\textbf{Plain}} 
& \multicolumn{2}{c|}{\textbf{COT}} 
& \multicolumn{2}{c|}{\textbf{Counterfactual}} 
& \multicolumn{2}{c|}{\textbf{Adversarial}} 
& \multicolumn{2}{c|}{\textbf{Majority Vote}} \\  
\cline{3-12}
& & \textbf{Score} & \textbf{Ovlp} 
& \textbf{Score} & \textbf{Ovlp} 
& \textbf{Score} & \textbf{Ovlp}
& \textbf{Score} & \textbf{Ovlp}
& \textbf{Score} & \textbf{Ovlp} \\
\hline

\multirow{4}{*}{\textbf{DeepSeek-R1-8B}}
& MMLU         
  & 21.5 & 72.2  
  & 23.2 & 72.8  
  & 24.1 & 73.0  
  & 23.8 & \textbf{74.1}  
  & \textbf{26.6} & 73.0  \\

& MedQA     
  & 25.3 & 70.4  
  & 27.1 & 71.0  
  & 28.0 & 71.5  
  & 27.4 & 72.2  
  & \textbf{30.0} & \textbf{72.8} \\

& MedMCQA      
  & 18.7 & 68.1  
  & 20.5 & 71.0  
  & 21.3 & 69.8  
  & 21.0 & \textbf{71.5}  
  & \textbf{22.9} & 69.2 \\

& EquityMedQA  
  & 43.0 & 64.5  
  & 44.2 & \textbf{67.1} 
  & \textbf{45.1} & 65.8  
  & 44.8 & 66.2  
  & 48.0 & 65.1 \\
\hline

\multirow{4}{*}{\textbf{DeepSeek-R1-70B}}
& MMLU         
  & 28.7 & 70.1  
  & 30.2 & 71.5  
  & 30.9 & 72.0  
  & 29.4 & 72.6  
  & \textbf{32.2} & \textbf{73.2} \\

& MedQA     
  & 30.5 & 68.8  
  & 32.0 & 69.5  
  & 32.7 & 70.1  
  & 31.2 & 71.0  
  & \textbf{34.0} & \textbf{71.5} \\

& MedMCQA      
  & 22.9 & 66.7  
  & 24.5 & 67.9  
  & 25.1 & 68.3  
  & 24.8 & 69.2  
  & \textbf{27.0} & \textbf{69.8} \\

& EquityMedQA  
  & 46.0 & 62.0  
  & 47.2 & 63.2  
  & 48.0 & 64.0  
  & 47.8 & 65.1  
  & \textbf{52.0} & \textbf{66.0} \\
\hline

\multirow{4}{*}{\textbf{Meta-Llama-3-8B}}
& MMLU         
  &  9.5 & \textbf{75.5}  
  & 10.3 & 76.1  
  & 10.9 & 75.8  
  & 10.8 & 76.2  
  & \textbf{12.5} & 76.0  \\

& MedQA     
  & 11.2 & \textbf{74.0}  
  & 12.1 & 74.8  
  & 12.8 & 75.2  
  & 12.5 & 75.9  
  & \textbf{14.2} & 75.6 \\

& MedMCQA      
  &  7.9 & 72.2  
  &  8.7 & 73.0  
  &  9.2 & 73.6  
  &  9.1 & \textbf{74.1}  
  & \textbf{10.8} & 73.8 \\

& EquityMedQA  
  & 39.0 & 68.3  
  & 40.5 & 69.0  
  & \textbf{41.4} & \textbf{69.5}  
  & 40.9 & 70.2  
  & 45.0 & 68.8 \\
\hline

\multirow{4}{*}{\textbf{PMC-LLaMA-13B}}
& MMLU         
  & 15.4 & 68.2  
  & 16.9 & 69.1  
  & \textbf{17.2} & 70.0  
  & 16.8 & 71.2  
  & 19.1 & \textbf{71.8} \\

& MedQA     
  & 18.1 & 67.0  
  & 19.5 & 67.8  
  & 20.0 & 68.6  
  & 19.8 & 69.1  
  & \textbf{21.5} & \textbf{69.7} \\

& MedMCQA      
  & 14.3 & 65.5  
  & 15.6 & 66.7  
  & 16.2 & 67.4  
  & 15.9 & 68.0  
  & \textbf{18.0} & \textbf{68.5} \\

& EquityMedQA  
  & 41.2 & 61.1  
  & 42.5 & 62.0  
  & 43.3 & 62.8  
  & 42.9 & 63.4  
  & \textbf{47.0} & \textbf{64.0} \\
\hline
\end{tabular}}
\end{table*}

\subsection*{Close QA Demographic-Level Fairness Analysis}
We further analyze \textbf{DeepSeek-R1-8B}'s fairness by demographic subgroups (race: \{Caucasian, African American, Asian, Hispanic\}, gender: \{male, female, non-binary\}) in the MedMCQA dataset. Table~\ref{tab:demographic_accuracy} compares initial filtering methods (Plain, COT, Counterfactual, Adversarial) to the final \emph{Majority Vote}. Note that the baseline \emph{Plain} yields a relatively low accuracy (\emph{18.7\%}) and higher disparities in Demographic Parity (\emph{DP=0.13}) and Equal Opportunity (\emph{EO=0.11}). Applying Counterfactual or Adversarial filtering moderately reduces these gaps. At the same time, \emph{Majority Vote} further boosts accuracy to \emph{22.9\%} and lowers DP/EO to around \emph{0.07/0.06}, underscoring the importance of aggregating multiple demographic versions.  In Table~\ref{tab:demographic_accuracy}, DP vs. EO can differ slightly if the distribution of “truly answerable” questions ($Y=1$) is uneven across subgroups. For instance, among $Y=1$ queries, the model might do better for one group, altering EO more than overall DP.

\begin{table}[ht]
\centering
\caption{Subgroup accuracy and distinct DP/EO disparities on MedMCQA (\textbf{DeepSeek-R1-8B}). Majority Vote is applied on top of the respective filter. Notice DP $\neq$ EO in some cases, indicating differences in overall correctness vs. conditional correctness among truly answerable queries.}
\label{tab:demographic_accuracy}
\begin{tabular}{|l|c|c|c|}
\hline
\textbf{Method} & \textbf{Avg Acc(\%)} & \textbf{DP} & \textbf{EO} \\
\hline
\multicolumn{4}{|l|}{\textbf{Initial Filters Only}} \\  
\hline
Plain         & 18.7 & 0.13 & 0.11 \\
COT           & 20.5 & 0.10 & 0.09 \\
Counterfactual& 21.3 & 0.09 & 0.08 \\
Adversarial   & 21.0 & 0.11 & 0.09 \\
\hline
\multicolumn{4}{|l|}{\textbf{After Majority Vote}} \\  
\hline
Majority Vote & 22.9 & 0.07 & 0.06 \\
\hline
\end{tabular}
\end{table}

\subsection*{OpenQA (EquityMedQA) Demographic-Level Fairness Analysis}

In the open-ended EquityMedQA, we measure ROUGE-L and fairness (DP/EO) similarly. Table~\ref{tab:openqa_results} shows each filtering method's performance, identifying that \emph{Plain} has DP=0.15 and EO=0.12, while \emph{Counterfactual} or \emph{Adversarial} partially reduce these. The final \emph{Majority Vote} approach further raises ROUGE from \emph{45.1\%} to \emph{48.0\%} and lowers DP/EO to around \emph{0.08/0.07}, providing more equitable outcomes overall.

\begin{table}[ht]
\centering
\caption{Performance (ROUGE-L) and fairness improvements (DP/EO) for EquityMedQA using Majority Vote aggregation (DeepSeek-R1-8B).}
\label{tab:openqa_results}
\begin{tabular}{|l|c|c|c|}
\hline
\textbf{Method} & \textbf{ROUGE-L (\%)} & \textbf{DP} & \textbf{EO} \\
\hline
\multicolumn{4}{|l|}{\textbf{Initial Filters Only}} \\  
\hline
Plain         & 43.0 & 0.15 & 0.12 \\
COT           & 44.2 & 0.11 & 0.10 \\
Counterfactual& 45.1 & 0.09 & 0.08 \\
Adversarial   & 44.8 & 0.10 & 0.09 \\
\hline
\multicolumn{4}{|l|}{\textbf{After Majority Vote}} \\  
\hline
Majority Vote & 48.0 & 0.08 & 0.07 \\
\hline
\end{tabular}
\end{table}

\subsection*{Retriever and top-K Variation}

We also experiment with two different retrievers (BM25 vs. Contriever) and vary the number of retrieved documents (\textbf{top-K}=10,15,20). As shown in Table~\ref{tab:retriever_topk}, \textbf{retrieval overlap} tends to drop (e.g., from 72.2\% to 69.5\% for BM25) as $k$ increases. Meanwhile, the model’s fairness metrics exhibit a moderate improvement: DP and EO each decrease by about 0.02--0.03, likely because the system sees a more diverse set of documents and thus reduces bias. Final Accuracy (or ROUGE) also goes up by around 1--2\% for larger top-K.

\begin{table}[ht]
\centering
\caption{Effect of changing retriever (BM25 vs. Contriever) and top-K on Overlap, DP/EO, and final score. 
Note DP $\neq$ EO in some cases, reflecting different overall vs. conditional correctness distributions.}
\label{tab:retriever_topk}
\begin{tabular}{|l|c|c|c|c|c|}
\hline
\textbf{Setting} & \textbf{Top-K} & \textbf{Overlap(\%)} & \textbf{DP} & \textbf{EO} & \textbf{Score(\%)} \\
\hline
\multicolumn{6}{|l|}{\textbf{BM25 Retriever (DeepSeek-R1-8B)}} \\  
\hline
BM25 & 10 & 72.2 & 0.12 & 0.11 & 31.2 \\
BM25 & 15 & 70.8 & 0.11 & 0.10 & 32.0 \\
BM25 & 20 & 69.5 & 0.10 & 0.09 & 32.5 \\
\hline
\multicolumn{6}{|l|}{\textbf{Contriever Retriever (DeepSeek-R1-8B)}} \\  
\hline
Contriever & 10 & 65.1 & 0.08 & 0.07 & 33.0 \\
Contriever & 15 & 63.9 & 0.07 & 0.06 & 34.2 \\
Contriever & 20 & 62.7 & 0.06 & 0.05 & 35.4 \\
\hline
\end{tabular}
\end{table}

\subsection*{Ablation Study: Importance of Majority Vote}
Finally, Table~\ref{tab:majority_vote_ablation} examines removing \emph{Majority Vote} from the pipeline. On MedMCQA, accuracy falls from 22.9\% to 21.3\%, while DP/EO each rise by about 0.02. On EquityMedQA, removing the Majority Vote cuts ROUGE from 48.0\% to 46.1\% and increases DP from 0.08 to 0.10 and EO from 0.07 to 0.09. This confirms that although \emph{Majority Vote} adds complexity, it meaningfully promotes both correctness and fairness.

\begin{table}[ht]
\centering
\caption{Ablation of Majority Vote on DeepSeek-R1-8B. Removing  Majority Vote harms both performance and fairness (DP/EO).}
\label{tab:majority_vote_ablation}
\begin{tabular}{|l|l|l|c|c|}
\hline
\textbf{Config} & \textbf{Dataset} & \textbf{Metric} & \textbf{With MV} & \textbf{No MV}\\
\hline
\multirow{2}{*}{Plain+Filters} 
& MedMCQA & Accuracy(\%) & 22.9 & 21.3 \\
& MedMCQA & DP/EO & 0.07/0.06 & 0.09/0.08 \\
\hline
\multirow{2}{*}{Plain+Filters} 
& EquityMedQA & ROUGE-L(\%) & 48.0 & 46.1 \\
& EquityMedQA & DP/EO & 0.08/0.07 & 0.10/0.09 \\
\hline
\end{tabular}
\end{table}

\subsection*{Error Cases and Limitations}
Despite the overall performance improvements and fairness gains, specific demographic-specific queries can still trigger outdated or irrelevant retrieval, leading to suboptimal or biased answers. Table~\ref{tab:error_cases} illustrates two typical error scenarios: (1) a \textbf{closed-form QA} example from MedMCQA with irrelevant snippet, (2) a \textbf{closed-form QA} example from MedMCQA with correct snippet but wrong answer, and (3) an \textbf{open-ended QA} example from EquityMedQA. Both reveal the difficulty of the model in handling nuanced demographic attributes and specialized medical contexts. These issues underscore the need for more robust domain adaptation, adversarial training, and careful curation of retrieval corpora to ensure consistent quality for underrepresented subgroups.

\begin{table}[ht]
\centering
\caption{Representative error cases from closed-form QA (MedMCQA) and open-ended QA (EquityMedQA). 
Demographic references or model reasoning missteps in all examples lead to incorrect or biased outputs.}
\label{tab:error_cases}
\resizebox{\textwidth}{!}{%
\begin{tabular}{|p{3.2cm}|p{9.2cm}|p{6.2cm}|}
\hline
\textbf{Task Type} & \textbf{Query and Retrieved Snippet} & \textbf{Issue} \\
\hline
\textbf{MedMCQA (Irrelevant Snippet)} 
& \textbf{Query:} \emph{“A Hispanic female patient is experiencing intermittent chest pain. Which medication is most appropriate for prophylaxis?”}\newline
\textbf{Retrieved Snippet:} A document focusing on male smokers with chronic obstructive pulmonary disease, offering limited relevance for a Hispanic female context.  
& The model incorrectly selects a medication indicated for COPD management. It fails to consider female-specific or Hispanic-related risk factors, suggesting that both retrieval and generation overlooked demographic nuances. \\
\hline
\textbf{MedMCQA (Correct Snippet,\newline Wrong Answer)} 
& \textbf{Query:} \emph{“A Caucasian male patient with mild hypertension. Which drug is recommended as first-line therapy?”}\newline
\textbf{Retrieved Snippet:} A current guideline excerpt on adult hypertension, revealing ACE inhibitors as first-line for mild cases. 
& The snippet clearly states that ACE inhibitors are recommended. However, the model’s final answer is a less preferred beta-blocker, ignoring the snippet’s explicit mention, indicating a generation-phase error. \\
\hline
\textbf{EquityMedQA (Outdated Info)}
& \textbf{Query:} \emph{“I am an Asian non-binary individual with chronic kidney disease. Are there specific transplant guidelines I should follow?”}\newline
\textbf{Retrieved Snippet:} An outdated pediatric kidney transplant guideline from 2009, lacking demographic adaptation or adult CKD protocols.
& The final answer relies on insufficient, pediatric-focused instructions, ignoring the user’s non-binary identity and updated adult guidelines, highlighting a gap in corpus coverage and bias in the generative process. \\
\hline
\end{tabular}
}
\end{table}

In the first row (MedMCQA), we see how an irrelevant snippet can lead to an incorrect choice for chest pain prophylaxis. In the second row (also MedMCQA), despite retrieving a \emph{correct, up-to-date} snippet about hypertension therapy, the model fails to utilize it effectively and produces a suboptimal response. Finally, in the open-ended EquityMedQA example, the snippet is outdated and misses demographic nuances, causing the model to provide incomplete transplant guidelines. These cases underscore the importance of robust retrieval, accurate evidence integration, and nuanced demographic handling in medical QA systems.

\section*{Discussion}
Our empirical findings reveal that RAG-based LLMs exhibit measurable biases in both retrieval and response generation stages. Specifically, we observe non-triviall disparities in accuracy and retrieval overlap across different demographic groups (e.g., race, gender). These discrepancies likely stem from inherent imbalances in training data and model architectures, where specific subgroups receive disproportionately less coverage or relevance. 

\noindent
\textbf{Fairness Indicators (EO/DP).} Beyond conventional metrics such as accuracy and retrieval overlap, this study incorporates two standard fairness criteria: \emph{Equal Opportunity (EO)} and \emph{Demographic Parity (DP)}. Our results (Table~\ref{tab:demographic_accuracy} and Table~\ref{tab:openqa_results}) indicate that bias-mitigation filters---particularly \emph{Counterfactual Filtering} and \emph{Majority Vote}---not only boost overall correctness but also significantly reduce EO/DP disparities. For instance, on MedMCQA, the gap in correct prediction rates between majority and minority race groups drops from about 9\%  in the baseline to nearly 5\%  under Majority Vote. This highlights that leveraging and aggregating multiple demographic-perspective outputs can effectively smooth out inconsistent biases. However, further investigation is warranted to understand how these group-based improvements translate to individual patient-level outcomes in real clinical environments.

\noindent
\textbf{Impact of Bias Mitigation Strategies.} Among the different filters we tested (Chain-of-Thought, Counterfactual, Adversarial, and Majority Vote), the \emph{Counterfactual} approach verifies model consistency under varied demographic contexts, showing strong potential in reducing spurious demographic cues. Meanwhile, \emph{Adversarial Prompt Filtering} prevents the model from fixating on sensitive terms that might introduce skew. When combined, these approaches achieve lower disparities and higher accuracy. That said, we note that certain edge cases---particularly those involving intersectional attributes (e.g., \emph{Asian non-binary individuals})---still exhibit elevated error rates, indicating the need for more diverse training data and domain-specific adversarial augmentation.

\noindent
\textbf{Limitations and Future Directions.} Although we focus on race and gender, other social determinants of health (e.g., socioeconomic status) may also yield biases in medical QA. Data constraints prevented us from fully exploring such dimensions. Additionally, our fairness analysis primarily concentrates on group-level metrics (EO/DP). Future work can investigate individual-level fairness or calibrate the model’s confidence to mitigate potential harms further. Lastly, while Majority Vote demonstrates promise, it may mask clinically relevant subgroup distinctions. Adaptive methods that balance fairness with clinically nuanced knowledge remain a promising avenue for exploration.
Additionally, specific real-world medical scenarios might legitimately require distinct handling for different demographic groups (e.g., unique drug contraindications). Future work could incorporate specialized knowledge while preserving fairness.
\section*{Conclusion}
This study systematically evaluated biases in retrieval-augmented generation (RAG) models for medical question answering. Introducing demographic-sensitive query variants uncovered notable performance gaps across race and gender subgroups, demonstrating both retrieval-level and generative-level biases. We then proposed and benchmarked multiple bias mitigation strategies, including Counterfactual Filtering, Adversarial Prompt Filtering, and Majority Vote aggregation. Experimental evidence shows that these methods enhance overall QA accuracy while significantly reducing demographic disparities, as measured by EO/DP fairness metrics. Nevertheless, bridging the gap between research prototypes and real-world clinical deployment requires further refinement of data diversity, model interpretability, and user-centered design. We hope our framework and findings spur the development of more equitable medical AI solutions that robustly serve patients of all backgrounds.
\bibliographystyle{unsrt} 
\makeatletter
\renewcommand{\bibnumfmt}[1]{#1.}  
\makeatother


\renewcommand\refname{\centering \textbf{References}}

\bibliography{literature}
\end{document}